# Aerial Image Stitching Using IMU Data from a UAV

Selim Ahmet IZ and Mustafa UNEL
*Faculty of Engineering and Natural Sciences*
*Sabanci University*
Istanbul, Turkey
{izselim, munel}@sabanciuniv.edu

*Abstract*— Unmanned Aerial Vehicles (UAVs) are widely used for aerial photography and remote sensing applications. One of the main challenges is to stitch together multiple images into a single high-resolution image that covers a large area. Feature-based image stitching algorithms are commonly used but can suffer from errors and ambiguities in feature detection and matching. To address this, several approaches have been proposed, including using bundle adjustment techniques or direct image alignment. In this paper, we present a novel method that uses a combination of IMU data and computer vision techniques for stitching images captured by a UAV. Our method involves several steps such as estimating the displacement and rotation of the UAV between consecutive images, correcting for perspective distortion, and computing a homography matrix. We then use a standard image stitching algorithm to align and blend the images together. Our proposed method leverages the additional information provided by the IMU data, corrects for various sources of distortion, and can be easily integrated into existing UAV workflows. Our experiments demonstrate the effectiveness and robustness of our method, outperforming some of the existing feature-based image stitching algorithms in terms of accuracy and reliability, particularly in challenging scenarios such as large displacements, rotations, and variations in camera pose.

## I. INTRODUCTION

Unmanned Aerial Vehicles (UAVs) have become increasingly popular for aerial photography and remote sensing applications. One of the main challenges in these applications is to stitch together multiple images captured by the UAV into a single high-resolution image that covers a large area. The stitching process involves aligning the images to a common coordinate system and blending them together to avoid visible seams, and this is an important problem for robotics applications [1, 2, 3].

There are two primary methods for image stitching: the Direct Method and Feature-based Methods. The Direct Method compares the pixel intensities of the overlapping sections to identify similarities between images. However, it is computationally complex, has limited convergence range, and is restricted by scale and rotation [4, 5]. On the other hand, Feature-based Methods, which include SIFT [6], SURF [7], FAST [8], ORB [9], and Harris Corner [10], detect the similarity of input images based on a few key feature points, resulting in faster results and more reliable output. However, these methods may not be suitable for images with a single color or entirely flat background, and the interpretation of loss data of input images in traditional feature-based techniques is limited.

Feature-based image stitching algorithms have been widely used for solving this problem. These algorithms rely on detecting and matching visual features in the images to estimate the relative positions and orientations of the cameras. However, these algorithms can suffer from errors and ambiguities in feature detection and matching, which can lead to incorrect alignment and visible seams in the stitched image [11]. To address these challenges, several approaches have been proposed in the literature.

One common approach is to use Bundle Adjustment [12] techniques to simultaneously estimate the camera poses and the 3D structure of the scene. This approach can achieve high accuracy and robustness, but it requires a large number of computational resources and can be computationally expensive. Another approach is Distortion Correction [13] which correct lens distortions in individual images before stitching to achieve better alignment, but it requires accurate calibration parameters, and not always effective in correcting severe distortions or complex lens aberrations. Moreover, the Random Sample Consensus (RANSAC) [14] is another feature-based stitching approach which estimates the transformation (translation, rotation, and scaling) between matched keypoints. However, RANSAC may not always find the optimal transformation, leading to stitching errors, and it is too sensitive to the presence of a high number of outliers or incorrect matches. Furthermore, Seam Finding and Blending technique [15] identifies the optimal seam (boundary) between overlapping images to ensure smooth transitions, and blends the images along the seams to minimize visible discontinuities, but blending can introduce artifacts like ghosting or color inconsistencies, and the approach does not work properly in complex scenes with moving objects or dynamic changes between images.

In this paper, we present a novel method for stitching images captured by a UAV using a combination of IMU data and computer vision techniques (see Fig. 1). Our method addresses several limitations of existing feature-based image stitching algorithms by exploiting the additional information provided by the IMU data. Specifically, we use the IMU data to estimate the displacement and rotation of the UAV between consecutive image captures, which provides a more accurate and reliable estimate of the camera poses. Our method involves several steps, including: (1) estimating the translation and rotation from the IMU data [16], (2) correcting for the skew perspective distortion [17] caused by the pitch angle of the camera and the altitude of the UAV, and (3) computing a homography matrix that maps the coordinates in the world frame to the coordinates in the camera frame.

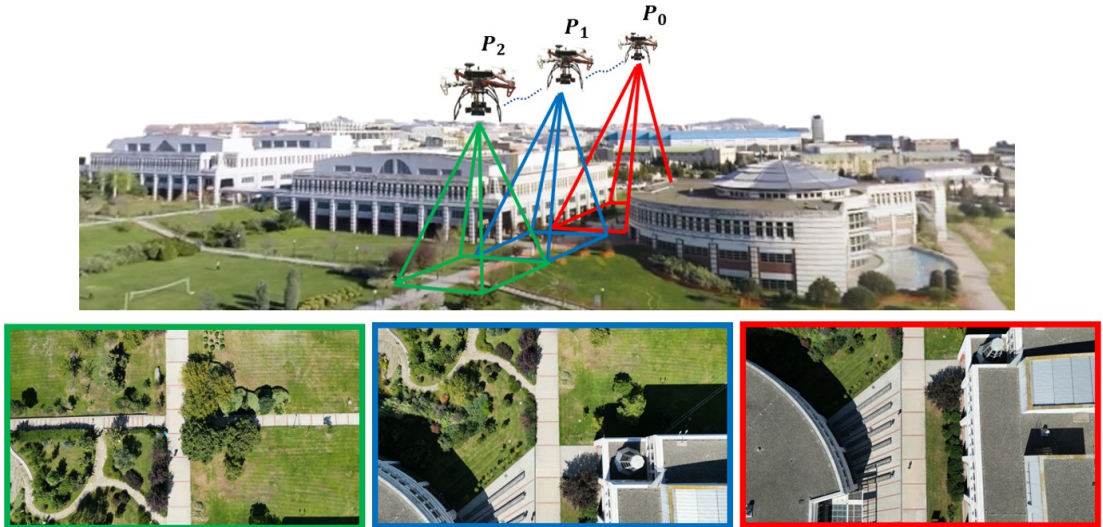

Fig. 1. Image stitching using onboard UAV data

Once the preprocessing is done, we use a standard image stitching algorithm to align and blend the images together. Our contributions in this paper can be summarized as follows:
- We propose a novel method for stitching images captured by a UAV that combines IMU data and computer vision techniques to achieve more accurate and reliable results.
- We demonstrate the effectiveness and robustness of our method on a dataset of aerial images captured by a quadrotor UAV, showing that it can successfully stitch together images with minimal visible seams and distortions.
- We compare our method with existing feature-based image stitching algorithms and show that it outperforms them in terms of accuracy and reliability, particularly in challenging scenarios such as large displacements, rotations, and variations in camera pose.

Our proposed method has several advantages over existing feature-based image stitching algorithms. Firstly, it leverages the additional information provided by the IMU data, which improves the accuracy and reliability of the camera pose estimation. Secondly, it corrects for various sources of distortion that can affect the quality of the stitched image, such as the skew perspective caused by the pitch angle and altitude of the camera. Finally, it can be easily integrated into existing UAV platforms and workflows, making it a practical and valuable tool for aerial photography and remote sensing applications.

The remainder of this paper is organized as follows: The Section II presents a detailed de- scription of the existing stitching methods and proposed approach including their advantages and limitations. The Section III provides the comparative analysis of the existing feature- based techniques and proposed algorithm in several scenarios. Finally, section IV concludes the paper and highlights the potential applications of these techniques besides possible future works.

## II. METHODOLOGY

There exist two primary approaches for image stitching, namely, feature-based methods and direct methods. Both direct and feature-based techniques are rooted in the identification of unique key points in input images, which are matched between the images. Subsequently, these techniques compute the homography matrix to map the corresponding reference points from one image to another and warp the input images into a common coordinate system. Finally, the warped images are blended together to create a visually continuous panorama. Notably, the key contrast between these methods lies in the feature detection and matching step. While feature-based methods rely on identifying and matching distinctive features between images, direct methods do not require feature detection and matching, instead using pixel intensities to align the images.

Direct methods are typically faster than feature-based methods because they do not rely on feature detection and matching. Nevertheless, direct methods have some drawbacks, including a complicated structure, a limited convergence range, and the inability to scale and rotate images [11]. In contrast, feature-based methods can handle larger image differences and non-rigid deformations by identifying and matching distinctive features between images. Matched features in feature-based methods can also be used to generate a panoramic mosaic with each feature mapped to its corresponding location in the input images, providing more detailed and interpretable results. A crucial step in the feature-based stitching method, which is not present in the direct method, is the use of RANSAC [18] to eliminate outliers and enhance the accuracy of feature matching (see Fig. 2). The direct stitching method suffers from certain constraints, such as a complex structure and a limited convergence range, which necessitate the use of feature-based stitching techniques during experiments to evaluate the efficacy of the proposed technique. The homograph estimation step in the global matching process allows the detected and matched feature pairs to be aligned.

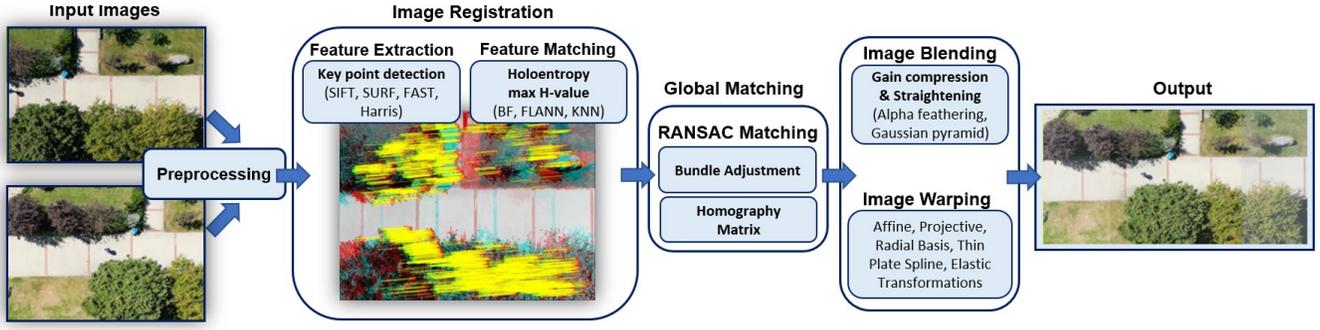

Fig. 2. Feature-based image stitching steps

One of the most straightforward approaches of computation of homography between input images can be expressed as:

$$H_{rn} = \prod_{i=r}^{n-1} H_{i(i+1)} \quad (1)$$

where $H_{01}$ to $H_{(n-2)(n-1)}$ are pairwise homography estimations between the new input image $I_n$ and the reference image $I_r$ [19]. Despite the apparent simplicity of this method, errors tend to accrue rapidly due to its multiplicative nature, causing the images to shift in the mosaic. This problem is minimized in Normalized Direct Linear Transformation (NDLT) algorithm [19] whose cost function can be represented as:

$$J(H_{i(i+1)}) = \|x_i - H_{i(i+1)} x_{i+1}\|^2 \quad (2)$$

However, this approach also has another drawback which is disappearing the minimum error property between the last 2 input images $I_j$ and $I_i$. There exists an alternative approach to overcome this problem as well. In that technique, the homography estimation is computed by using a new image and the reference image which is already stitched in the previous step. The cost function of this approach is written as:

$$J(H_{ri}) = \|H_{r(i-1)} x_{i-1} - H_{ri} x_i\|^2 \quad (3)$$

where $x_i$ and $x_{i-1}$ are the matching features [19]. Following the completion of the previous steps, a series of post-processing operations, including image blending and image warping, are executed to enhance the accuracy and visual quality of the resulting output. Image blending is the process of smoothly merging the overlapping portions of stitched images to create a seamless and visually appealing panorama. Conversely, image warping is the practice of modifying stitched images to account for geometric distortions such as lens distortion or perspective distortion [11, 20].

Both direct and feature-based techniques encounter common challenges such as being unable to produce robust results when the overlapping regions of input images are small or when the main theme of the input images is quite similar. In such cases, algorithms may struggle to identify common features or verify pixel intensities due to the limited common region, or the high level of similarity between the input images. The proposed method entails the determination of relative positions of images obtained from the IMU records followed by image stitching. This method does not rely on the extraction of image features, decoding of pixel intensities, or investigation of image similarity, as in the feature-based and direct methods, respectively. Thus, the efficacy of the proposed technique must be evaluated independently of these two primary approaches.

*A. The Proposed Aerial Image Stitching Algorithm*

The proposed method was evaluated using a quadrotor equipped with an IMU and a monocular camera. The quadrotor captured high-resolution images of rough terrain at altitudes ranging from 15 to 25 meters. The IMU provided acceleration and orientation data, allowing for the computation of the UAV's location. A marker at the take-off station helped identify the initial position and orientation. An ultrasonic sensor measured the quadrotor's altitude. Position and altitude information from the onboard sensors, along with the captured images, were used to determine displacement. The images were then shifted and stacked using the calibrated onboard camera, converting the displacement to pixel values based on the camera's known parameters.

Double integration of acceleration with respect to time is required to determine displacement from IMU data. However, the integration process accumulates errors, known as drift, due to the presence of noise in the acquired data. To tackle this issue, numerical methods like the Trapezoidal rule and Simpson's rule are crucial. These methods enable incremental displacement computation by maintaining a uniform sampling rate from the IMU, which helps mitigate drift.

After the displacement of the quadrotor when it takes the images are computed, it is necessary to convert the metric data to pixel values to offset and stitch the new captured image to top of the old one. In this step, we can use this displacement vector to stitch the images together (see Fig. 3). One way to do this is to apply a projective transformation to one of the images such that it aligns with the other image. Let's suppose that we want to align image $I_k$ and $I_{k+1}$. We can use the following projective transformation matrix:

$$Q_{k,k+1} = [R_k | t_k] \quad (4)$$

where $R_k$ is a 3x3 rotation matrix that aligns the two images, and $t_k$ is the 3x1 translation vector that accounts for the displacement between the two positions. The matrix $Q_{k,k+1}$ maps the coordinates in image $I_k$ to the coordinate in image $I_{k+1}$.

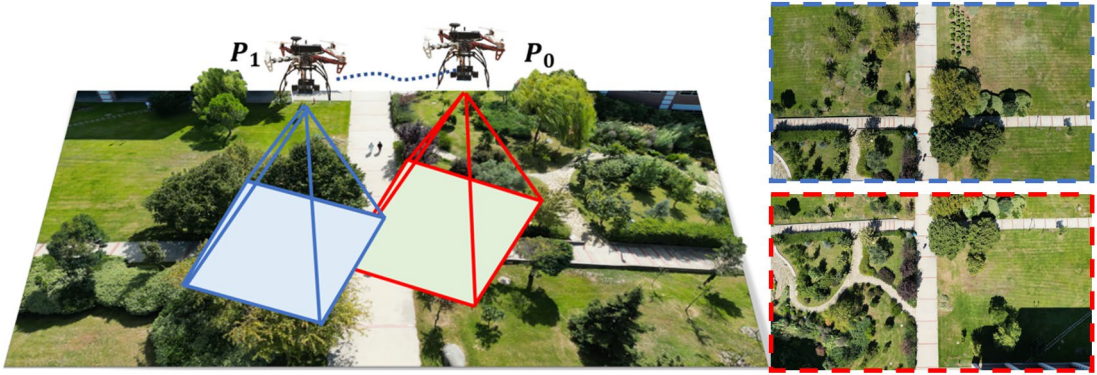

Fig. 3. Images captured at two different configurations

To apply this transformation to the image $I_k$ we first need to define a reference frame for the image. Let $O_k$ be the origin of the reference frame, and let $u_k$ and $v_k$ be the unit vectors that define the x and y axes, respectively. We can choose these vectors such that they correspond to the horizontal and vertical directions in the image. Next, we define the coordinates of the four corners of the image in the reference frame:

$$p_1 = [0\ 0\ 1] \quad p_3 = [m\ n\ 1]$$
$$p_2 = [m\ 0\ 1] \quad p_4 = [0\ h\ 1] \tag{5}$$

where $m$ and $n$ are the width and the height of the reference mage, respectively. We can then apply the transformation matrix $Q_{k,k+1}$ to these coordinates to obtain their positions in the reference frame of the image $I_{k+1}$:

$$q_{1,\ldots,4} = Q_{k,k+1} p_{1,\ldots,4} \tag{6}$$

Finally, we can compute the transformation matrix that maps the coordinates in the reference frame of image $I_k$ to the coordinates in the reference frame of image $I_{k+1}$:

$$T_{k,k+1} = M_{k+1} Q_{k,k+1} M_k^{-1} \tag{7}$$

where $M_k$ and $M_{k+1}$ are the matrices that define the affine transformation between the reference frame of each image and the image plane. These matrices can be calculated based on the position and orientation of the UAV and the focal length of the camera. Finally, we can apply the transformation matrix $T_{k,k+1}$ to the image $I_{k+1}$ to obtain the aligned image with $I_k$.

*1) Altitude Difference Effect*

When the UAV changes altitude between two consecutive image captures, the scale of the images change. To correct for this effect, we need to rescale the images so that they have the same scale.

Let's assume that the UAV flies at two different altitudes $h_k$ and $h_{k+1}$ when it captures the images at moments $t_k$ and $t_{k+1}$, respectively. The focal length of the camera is assumed to be constant. The scale factor $s_{k,k+1}$ between the two images can be calculated as follows:

$$s_{k,k+1} = \frac{h_{k+1}}{h_k} \frac{f}{f_k} \tag{8}$$

where $f$ is the focal length of the camera, and $f_k$ is the intrinsic parameter of the camera matrix of the image captured at moment $t_k$. The factor $(f/f_k)$ accounts for the difference in the intrinsic parameters between the two images.

To rescale the image captured at moment $t_k$, we can multiply the camera matrix by the scale factor $s_{k,k+1}$:

$$K'_k = s_{k,k+1} K_k \tag{9}$$

where $K_k$ is the original camera matrix of the image captured at moment $t_k$, and $K'_k$ is the new camera matrix after rescaling. We can then apply the rescaled camera matrix to the world points to obtain the corrected image points $p'_k$:

$$p'_k = K'_k Q_{k,k+1} P_k \tag{10}$$

where $[R_k | t_k]$ is the transformation matrix that maps the coordinates in the world frame to the coordinates in the camera frame at moment $t_k$, and $P$ is the 3D point in the world frame. As a final step, we need to perform this rescaling operation for each pair of consecutive images to ensure that all images have the same scale.

*2) Yaw Axis Motion Effect*

When the UAV changes its heading angle between two consecutive image captures, the captured images need to be rotated to align them properly. We can use the yaw angle measured by the UAV's IMU to calculate the rotation required. Let's assume that the UAV's heading angle is $yaw_k$ and $yaw_{k+1}$ when it captures the images at moments $t_k$ and $t_{k+1}$, respectively. We need to rotate the image captured at moment $t_k$ by an angle $\Delta_{yaw}$ around the z-axis to align it with the image captured at moment $t_{k+1}$. The value of $\Delta_{yaw}$ can be found as:

$$\Delta_{yaw} = yaw_{k+1} - yaw_k \tag{11}$$

To rotate the image captured at moment $t_k$ we can use a rotation matrix $R_z$ around the z-axis:

$$R_z = \begin{bmatrix} \cos(\Delta_{yaw}) & -\sin(\Delta_{yaw}) & 0 \\ \sin(\Delta_{yaw}) & \cos(\Delta_{yaw}) & 0 \\ 0 & 0 & 1 \end{bmatrix} \tag{12}$$

We can apply the rotation matrix $R_z$ to the camera matrix of the image to obtain the corrected camera matrix $K'_k$:

$$K'_k = K_k R_z \tag{13}$$

We can then use the corrected camera matrix $K'_k$ to transform the image to align it with the image captured at moment $t_{k+1}$:

$$p'_k = K'_k Q_{k,k+1} P_k \tag{14}$$

where $Q_{k,k+1}$ is the transformation matrix that maps the coordinates in the world frame to the coordinates in the camera frame at moment $t_k$, and $P$ is the 3D point in the world frame.

### 3) Pitch Axis Motion Effect

When the UAV captures images while in motion, the camera view may be skewed due to the pitch angle. This skew perspective transformation needs to be corrected before stitching the images. We can use the pitch angle measured by the UAV's IMU to calculate the transformation required.

Let's assume that the pitch angle of the UAV's camera is $\mu$ when it captures the image at moment $t$. We need to correct the skew perspective transformation by applying a shear transformation to the image. The shear transformation matrix S can be calculated as:

$$S = \begin{bmatrix} 1 & 0 & 0 \\ \tan(\mu) & 1 & 0 \\ 0 & 0 & 1 \end{bmatrix} \quad (15)$$

We can apply the shear transformation matrix S to the camera matrix of the image to obtain the corrected camera matrix K':

$$K' = K\,S \quad (16)$$

We can then use the corrected camera matrix K' to transform the image to correct the skew perspective:

$$p_k' = K'\, Q_{k,k+1}\, P_k \quad (17)$$

where $Q_{k,k+1}$ is the transformation matrix that maps the coordinates in the world frame to the coordinates in the camera frame, and $P_k$ is the 3D point in the world frame.

## III. RESULTS AND DISCUSSION

As a result, the proposed algorithm is developed in MATLAB environment and implemented various cases. Its performance is compared with the feature-based stitching algorithm performances in two different scenes. As it can be seen in the Fig. 1, the taken aerial images at the $P_0$, $P_1$, and $P_3$ points are stitched by following the proposed steps and the following result is obtained:

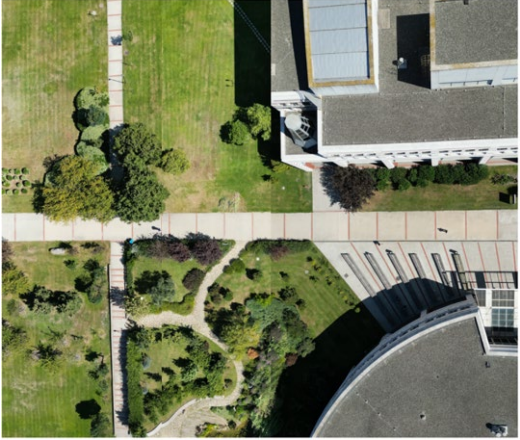

Fig. 4. Resulting mosaicked image

As it can be seen from the Fig. 4, the proposed algorithm stitches the target and reference images successfully. In order to see better the efficiency of the proposed technique, the feature-based techniques are also implemented into same input images.

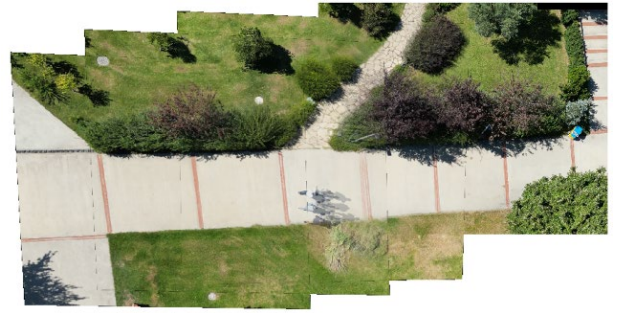

Fig. 5. Result of feature-based technique for flight scenario 1

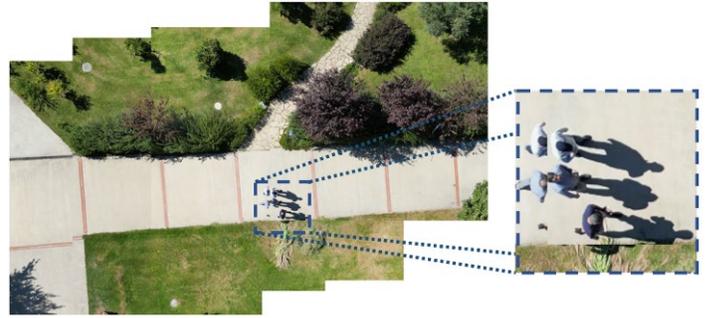

Fig. 6. Result of the proposed technique for flight scenario 1

As depicted in Fig. 5 and Fig. 6, it can be observed that the feature-based image stitching approach produced suboptimal results in terms of image quality during post-processing, resulting in blurry and degraded transitions between input images. Moreover, both the feature-based and proposed methods encountered challenges in addressing parallax error and visual distortion caused by object movement in the scene. Subsequently, both methods are evaluated for stitching two input images with similar backgrounds. While the feature-based approach identified key features in the input images, it failed to perform the match operation due to the similarities between these features. In contrast, the proposed method successfully computed the relative position of the image centers and seamlessly combined the target image with the reference image, demonstrating its superior performance in handling challenging scenarios (see Fig. 7 and Fig. 8)

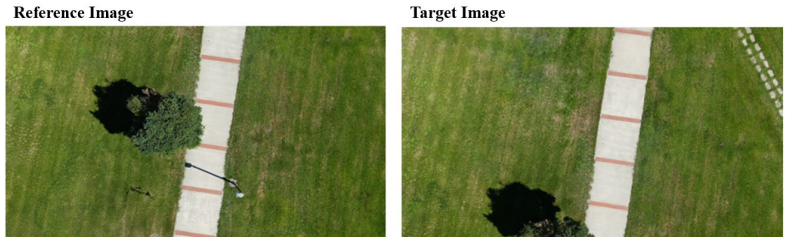

Fig. 7. Input images have similar background

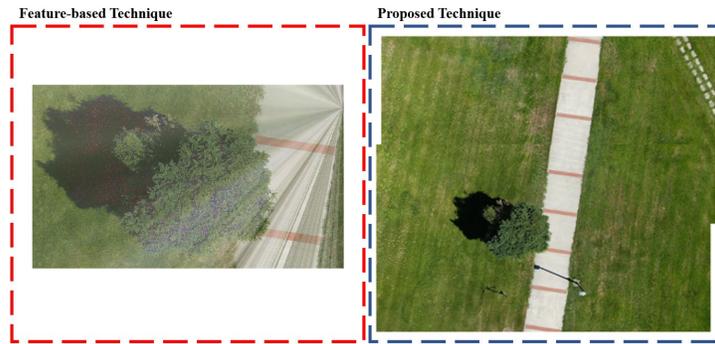

Fig. 8.  Feature-based (left) stitching and proposed technique (right) where the inputs have similar background

## IV. Conclusion

In conclusion, our proposed method for image stitching using IMU data and computer vision techniques provides a promising solution to overcome the limitations of feature-based algorithms. By leveraging the additional information provided by IMU data, we improve the accuracy and reliability of camera pose estimation, which is crucial for stitching multiple aerial images into a seamless panorama. Our method also addresses various sources of distortion, such as perspective and skew distortions caused by the camera's pitch angle and altitude, resulting in higher-quality stitched images. Moreover, our method is easily integrable into existing UAV platforms and workflows, making it a practical tool for aerial photography and remote sensing applications.

Overall, our experiments demonstrate the effectiveness and robustness of our method on a dataset of aerial images captured by a quadrotor UAV. Compared to some of the existing feature-based algorithms, our proposed method outperforms them in challenging scenarios with large displacements, rotations, and variations in camera pose. Therefore, we believe that our method can make a significant contribution to the field of computer vision and remote sensing, providing a more accurate and reliable approach to stitching aerial images.